\def\year{2022}\relax
\documentclass[letterpaper]{article} 
\usepackage{aaai22}  
\usepackage{times}  
\usepackage{helvet}  
\usepackage{courier}  
\usepackage[hyphens]{url}  
\usepackage{graphicx} 
\usepackage{natbib}  
\usepackage{caption} 
\DeclareCaptionStyle{ruled}{labelfont=normalfont,labelsep=colon,strut=off} 
\usepackage{algorithm}
\usepackage{algorithmic}

\usepackage{newfloat}
\usepackage{listings}
\floatstyle{ruled}
\newfloat{listing}{tb}{lst}{}
\floatname{listing}{Listing}

\usepackage{xspace}
\usepackage{amsfonts}
\usepackage{booktabs}
\usepackage{textcomp}
\usepackage{gensymb}
\usepackage{tabularx}
\usepackage[export]{adjustbox}
\usepackage{subfig}
\usepackage{xcolor}

\newcommand{\RNum}[1]{\uppercase\expandafter{\romannumeral #1\relax}}

\usepackage{mathtools}

\DeclarePairedDelimiterX{\norm}[1]{\lVert}{\rVert}{#1}
\DeclareMathOperator*{\argmin}{\arg\!\min}

\DeclarePairedDelimiter\floor{\lfloor}{\rfloor}

\newcommand{\hide}[1]{}

\newcommand{\CGDn}{Confidence Guided Distance Network\xspace}
\newcommand{\CGDnA}{CGD-net\xspace}
\newcommand{\CGD}{Confidence Guided Distance\xspace}
\newcommand{\CGDA}{CGD\xspace}

\newcommand{\xri}{\mathcal{X}_{RI}}
\newcommand{\yri}{\mathcal{Y}_{RI}}
\newcommand{\neigh}{\mathcal{N}_{x_i}}
\newcommand{\softP}{\mathcal{P}}

\newcommand{\source}{\mathcal{X}}
\newcommand{\target}{\mathcal{Y}}
\newcommand{\trans}{\mathcal{T}}

\title{Deep Confidence Guided Distance \\ for 3D Partial Shape Registration}

\author {
    Dvir Ginzburg,
    Dan Raviv
}
\affiliations {
    Tel-Aviv University\\
    dvirginzburg@mail.tau.ac.il, darav@tauex.tau.ac.il
}

\begin{document}

\maketitle

\begin{abstract}
We present a novel non-iterative learnable method for \textbf{partial-to-partial} 3D shape registration.
The partial alignment task is extremely complex, as it jointly tries to match between points, and identify which points do not appear in the corresponding shape, causing the solution to be non-unique and ill-posed in most cases. 

Until now, two main methodologies have been suggested to solve this problem: sample a subset of points that are likely to have correspondences, or perform soft alignment between the point clouds and try to avoid a match to an occluded part. These heuristics work when the partiality is mild or when the transformation is small but fails for severe occlusions, or when outliers are present.
We present a unique approach named \CGDn (\CGDnA), where we fuse learnable similarity between point embeddings and spatial distance between point clouds, inducing an optimized solution for the overlapping points while ignoring parts that only appear in one of the shapes.
The point feature generation is done by a self-supervised architecture that repels far points to have different embeddings, therefore succeeds to align partial views of shapes, even with excessive internal symmetries, or acute rotations.
We compare our network to recently presented learning-based and axiomatic methods and report a fundamental boost in performance.
\end{abstract}

\begin{figure}[!ht]
\centering

\setlength\tabcolsep{0pt}

\includegraphics[width=0.35\textwidth]{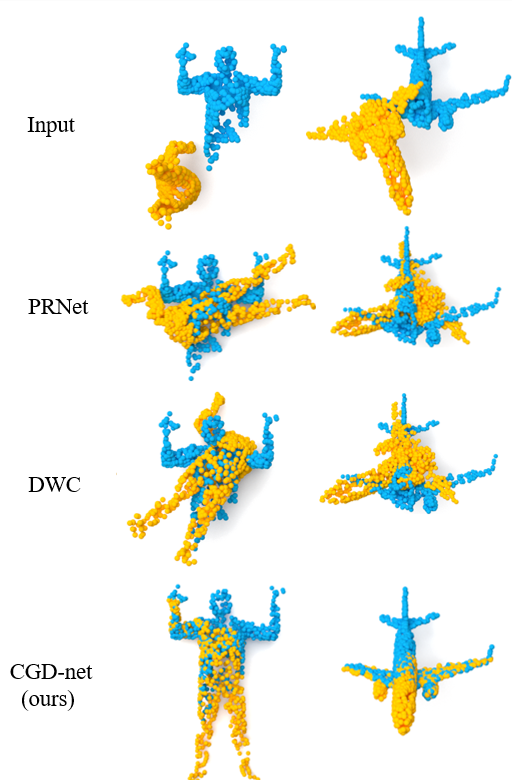}

\caption{Partial-to-Partial rigid alignment in the full spectrum of the rotation group SO(3) (Blue - source shape, Yellow - target shape). While other state-of-the-art methods struggle to align under such acute rotations and severe partiality, \CGDnA succeeds.
} \label{fig:tizzer_figure}

\end{figure}

\section{Introduction}

Shape registration is essential for a wide range of applications in computer vision, being the back-bone mechanism in numerous tasks, such as medical imaging \cite{medical_registration}, autonomous driving ~\cite{autonomous_registration}, and robotics ~\cite{robotics_registration}. 

A canonical work in the field is the Iterative Closest Point (ICP) algorithm \cite{icp},  which aligns shapes in iterations until convergence, by matching points according to spatial proximity, and then, solving a least-squares problem for the global transformation \cite{rtsolve}. 
ICP is extremely sensitive to noise and initial conditions and tends to get stuck in sub-optimal solutions \cite{icpl4}.
To overcome ICP's drawbacks, Deep Closest Point (DCP) \cite{dcp} replaced the Euclidean nearest point step of ICP with a learnable per-point embedding network, followed by a high-dimensional feature-matching stage.

In the partial registration problem, the task is to align partially overlapping segments of a canonical shape.
Recent methods have suggested choosing a subset of points as key points based on top-k heuristics and solving the optimal transformation between these essence points to address the partiality problem \cite{prnet,robustpointcloud}.
A crucial problem with partial 3D shape registration arises from the internal symmetries within the objects. Many point clouds are characterized by co-occurring segments, like the legs of a chair or an airplane's wings. 
When different co-occurring segments appear in the source and target shapes, these algorithms might still match them, resulting in wrong registrations.
Until now, only a few methods have been suggested for the partiality problem in a way that addresses the internal symmetries \text{and} avoids the hard selection of corresponding points.

To solve the above issues and provide a reliable yet self-supervised measure for the similarity of partial shapes, we present \CGD (\CGDA). \CGDA is a learning-based measure that coalesces points' spatial and latent similarity to a unified score by factoring the Euclidean distance between points with the confidence of their match by a deep neural network.
The architecture, named \CGDnA, constrains pooled clusters to span the embedding space, causing distant segments to have different embeddings, thus overcoming the internal symmetries problem of previous algorithms. 

We analyze the model's performance under popular point-cloud datasets and present superior results by a large margin compared to known state-of-the-art models in all datasets and configurations.

\paragraph{Contributions} We summarize our key contributions as follows:

\begin{itemize} 
\itemsep0em 
\item Introduce \CGD, a learnable measure that melds the spatial and latent proximity of points, to overcome previous metrics disadvantages for partial shape correspondence.
\item Present a new learnable paradigm for 3D partial shape registration that meets internal symmetries and severe partiality challenges.
\item Offer a non-iterative  network, both in training and test, granting the method robustness to initial conditions and high noise values.
\item Report state of the art results on a variety of synthetic and real-world datasets, while being robust to noise, sparsity, and extreme rotations.
\end{itemize}

\section{Related Work}

\paragraph{Axiomatic  3D shape  registration}
Point cloud registration has been researched for more than 40 years \cite{icp,chen1992object}, offering new capabilities in a wide range of computer vision applications. Iterative Closest Point (ICP) \cite{icp}, one of the seminal works in the field presented an iterative solution to the problem, by alternating between matching points via a simple nearest-neighbor heuristic and calculating the transformation that best describes the above match. While being a cornerstone for further research, ICP is extremely sensitive to noise, outliers, and initial conditions, and thus prone to reach a local-minima.
Many variants of ICP have been proposed to solve the caveats in the original solution \cite{icpl3,icpl4}. Go-ICP \cite{icpl1} offers an outlier detector scheme, while \citet{icpv1} formulates the registration error as the objective function of a non-linear
optimization problem (the Levenberg–Marquardt algorithm), refining the alignment until convergence. 
Still, none of the above methods is suitable for the partial-to-partial framework, and can only handle full-to-full or partial-to-full settings.
An important work to ours is ICP Registration Using Invariant Features \cite{sharp2002icp} which offered a fusion of Euclidean and rotation invariant features as part of the ICP process. It was also among the first to identify that rotation invariant features are less descriptive than the euclidean counterparts by construction, requiring the algorithm to find other information extraction pipelines, as we offer here.
\paragraph{Deep 3D shape registration}
In recent years, many works have tried to solve the 3D shape registration problem by harvesting Graph Neural Networks (GNNs) \cite{gnns1} capabilities to create descriptive per-point features. Research in the area is divided into two main categories, dense-shape correspondence \cite{cyclic, unsupmultispectral,dpc} and rigid alignment \cite{dcp,dwc,deepglobal}. While dense-shape correspondence is a vital research track, most real-world problems today deal with high-noise values, partiality, and outliers, in multi-sensor environments, requiring us to align the views to a unified scene.
Deep Closest Point (DCP) \cite{dcp} was among the first to propose a feature learning scheme, followed by a soft alignment of the two-point clouds instead of the spatial matching step presented in ICP. As DCP uses all points in the soft alignment step, it is sensitive to outliers or noise and does not work well for the partiality problem. 
Methods that follow DCP \cite{itnet,pointnetlk} have tried to enhance DCP's results using an iterative scheme similar to ICP. As with ICP, robustness to outliers and initial conditions remains an unsolved issue.
PRNet \cite{prnet} chooses a subset of points and solves the alignment only for the limited set to overcome the problem of outliers and partiality. 
Deep Weighted Consensus (DWC) \cite{dwc} suggests a different selection strategy, where a sampling distribution is defined based on the confidence of each source point in its alignment. DWC offers multiple possible transformations in parallel, and chooses the best parameter based on the Chamfer Distance \cite{chamfer} between the source and the transformed target. DWC is irrelevant for partial-to-partial alignment, as choosing based on the minimal chamfer distance between partial samples leads to results that are far from the true transformation, as exemplified in figure \ref{fig:tizzer_figure}.

\paragraph{Partial registration} 
The partial registration problem is the most relevant for real-world scenarios, thus became extremely researched in recent years \cite{deepglobal,robustpointcloud,DG2N}. As mentioned, PRNet \cite{prnet} offered a hard sampling heuristic to choose the points subset the are present both in the source and target point clouds and compute the transformation only between the subsets.
Other methods as \cite{predator,3d3d} offered attention mechanisms that inspect both clouds jointly, weighting points with high probability to appear in both partial shapes. While surpassing previous methods, co-occur segments and high noise values were still a problem due to local feature generation and reliance on the initial state of the point clouds.
Another interesting concept was introduced in \cite{dec2,dec3} where instead of sampling relevant points, a decoder network generated a full shape from the partial scans, offering a method to overcome the partiality problem. While this is a novel line of work, it is less relevant for generalization tasks, where we train on one dataset and infer on another.

\section{Deep Confidence Guided Distance}\label{sec:DCGD}

Given a transformation $\trans$, a source shape $\source$, and a target $\target$, the ability to define a distance measure $d(\trans(\source),\target)$ that is invariant to outliers and non-overlapping segments is crucial for the success of partial-to-partial algorithms.
Chamfer distance (CD) \cite{chamfer}, a popular distance measure between 3D objects, computes the distance from each source point to its target shape nearest neighbor, and vice-versa. 
We denote $nn_{\mathcal{C}}(p)$ as the euclidean closest point to $p$ in a point cloud $\mathcal{C}$, and $d_{nn}(p,\mathcal{C})$ to be that euclidean distance.
Formally, the metric takes the following form:
\begin{align} \label{eq:naive_chamfer}
CD(\trans(\source),\target)= \sum\limits_{i=1}^{i=|\bar{\mathcal{X}}|} d_{nn}(\mathcal{T}(x_i),\mathcal{Y}) + \\ \nonumber \sum\limits_{j=1}^{j=|\bar{\mathcal{Y}}|} d_{nn}(y_i,\mathcal{T}(\source)) \\ 
\nonumber \text{where}\quad d_{nn}(p,\mathcal{C})=||p-nn_{\mathcal{C}}(p)||_2^2\quad &\text{and }\\ \nonumber \quad nn_{\mathcal{C}}(p) = \argmin_{c_i\in \mathcal{C}} ||p-c_i||_2^2.
\end{align}
With $\bar{\mathcal{X}}$ being the set of points that uphold $d_{nn}(x_i,\mathcal{Y})<2d_s$ where $d_s$ is the maximal sampling distance of the point clouds, as matches with higher $d_{nn}$ are considered as outliers.
Equation \ref{eq:naive_chamfer} fails for partial-to-partial alignment, or when the number of outliers is high. This is because CD is optimal in expectation, whereas for the partiality setting, one should strive to ignore segments that do not have a correspondence in the target point cloud. The problem of CD with partiality is discussed extensively in the literature \cite{badchamf1} and exemplified in the results section (Sec. \ref{sec:results}). Deep neural networks have the ability to represent points in contextualized latent spaces influenced by the shape topology and local geometry. Given such deep embeddings $\hat{h}_{x_i},\hat{h}_{y_j}$ of points $x_i,y_j$, a natural choice for the latent similarity of the points is the cosine similarity:
\begin{equation} \label{eq:cos}
  \cos(x_i,y_j)=\frac{ \hat{h}_{x_i} \cdot \hat{h}_{y_j}} {|| \hat{h}_{x_i} ||_2 \cdot ||\hat{h}_{y_j}||_2}.
\end{equation}
For compactness, we denote by  $\softP\in\mathbb{R}^{|\source|\times|\target|}$ the soft alignment mapping, where \begin{equation}\label{eq:softP}
\softP_{i,j} = \cos(x_i,y_j).
\end{equation}
Intuitively, given the per-point embeddings $\hat{h}_{x_i}$ and $\hat{h}_{y_j}$, $\softP_{i,j}$ represents the latent similarity between $x_i$ and $y_j$.
We propose \CGD (\CGDA), a learning-based measure that fuses together the spatial and latent proximity, in a way that penalizes point clouds that are close under the Chamfer distance metric but are highly dissimilar in the latent space. \CGD is formulated as:

\begin{align}
CGD(\trans(\source),\target)&=
\sum\limits_{i=1}^{i=|\bar{\mathcal{X}}|} exp(-\gamma\softP_{i,q})d_{nn}(\trans(x_i),\target) \\[1pt]\nonumber  + \sum\limits_{j=1}^{j=|\bar{\mathcal{Y}}|} &exp(-\gamma\softP_{r,j}) d_{nn}(y_j,\trans(\source)) \\ \nonumber
\text{with }q=nn_{\target}&(\trans(x_i)),r=nn_{\trans(\source)}(y_j).
\end{align}

where $\gamma$ is a scalar that determines the weight given to the latent similarity and is conditioned by the shape's scale and sampling density.
For high $\softP_{ij}$, the combined measure is lower, while for matches with low or negative $\softP_{ij}$ the distance increases. 
   
In the ablation study (Sec. \ref{subsec:ablation}) we provide analysis on the importance of the \CGDA metric and the implications of using Chamfer Distance as the consensus metric instead.

\section{Architecture}\label{sec:architecture}

The following section outlines the architecture building blocks needed for the evaluation of the \CGDA measure.

\begin{figure*}[t!]
  \centering
  
  \includegraphics[width=0.90\linewidth]{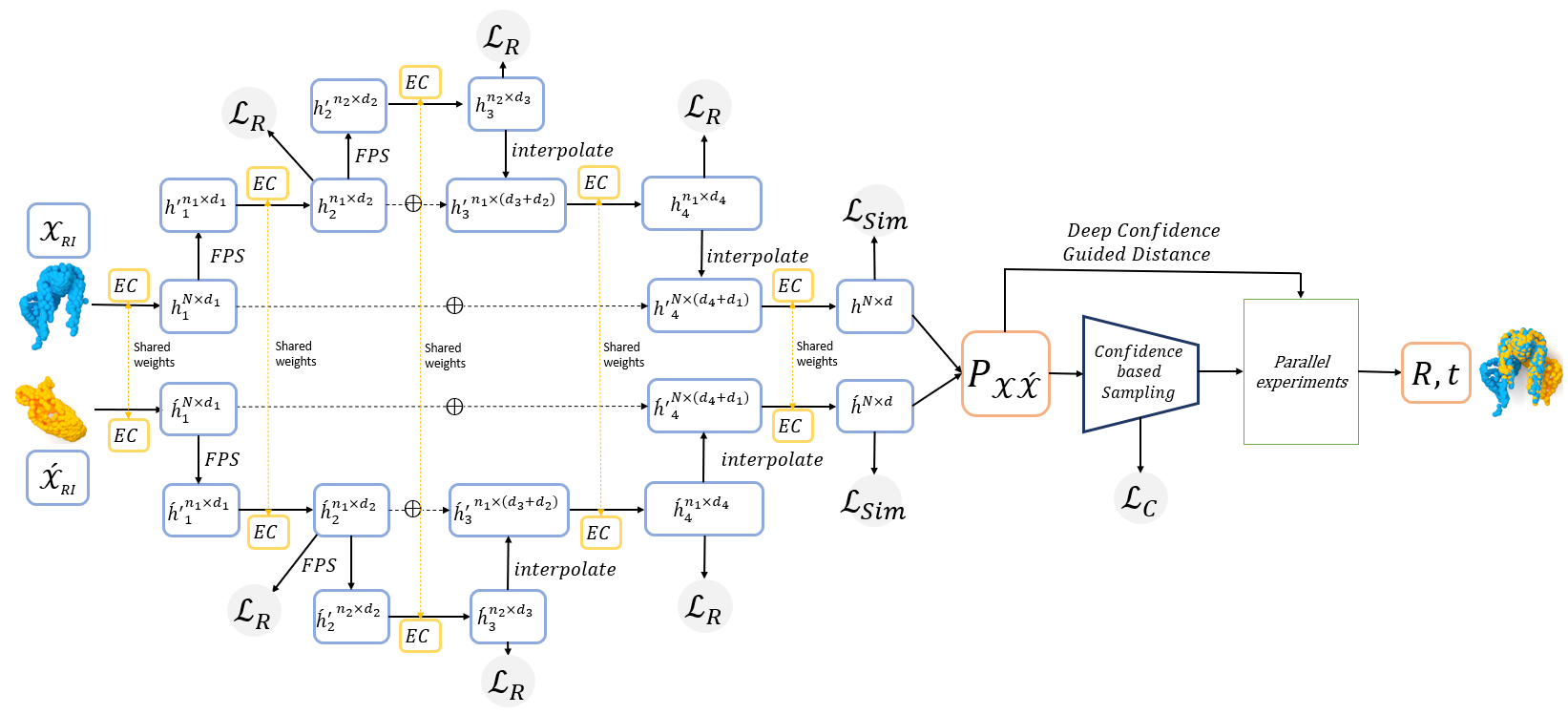}
  \caption{\CGDnA is comprised of three steps: (a)  Embed points via the hierarchical feature extraction network (Sec. \ref{subsec:feature_extraction}). (b)  Construct the confidence based sampling distribution of source points, formed from the soft correspondence matrix $\softP_{\mathcal{X}\mathcal{\grave{X}}}$ (Sec. \ref{subsec:softcor}). (c) Conduct multiple parallel experiments, and choose the experiment achieving the lowest \CGDA (Sec. \ref{sec:DCGD}).
  RI are the rotation invariant features, EC stands for EdgeConv \cite{dgcnn}, and FPS \cite{fps} refers to Euclidean farthest point sampling. $\mathcal{L_R},\mathcal{L}_{Sim}$ are the per-point repulsion and similarity loss functions, respectively, $\mathcal{L_C}$ is the self-supervised contrastive loss  (section \ref{subsec:loss}). The plus sign stands for concatenation.
}
  \label{fig:arch}

\end{figure*}

\subsection{Hierarchical feature extraction}
\label{subsec:feature_extraction}

\CGDn (\CGDnA)  presents a hierarchical graph neural network, inspired by PointNet++ \cite{pointnetpp} for intrinsic per-point feature extraction. The decision to use a hierarchical network arises from the repulsion loss (Sec. \ref{subsec:loss}) that pushes spatially far segments to have unique embeddings. As for the input point cloud representation, we adopt the rotation invariant (RI) features presented in DWC. 
Given rotation invariant features for the source and target shapes $\xri,\yri$, the input topology is defined by the $k$ nearest neighbors of each point. \CGDnA uses EdgeConv \cite{dgcnn} as the graph convolution operation, instead of PointNet++ suggested convolution, as PointNet operator is heavily biased toward spatial similarity, making it impractical for large rotations or high noise values. In EdgeConv, for a point $x_i$ and its neighborhood $\neigh$, the output embedding of $x_i$ is:
\begin{equation}
    h_{x_i}= \max\limits_{x_j \in \neigh} f_h([x_i||x_j]),
\end{equation}
where $f_h$ is a learnable function, $||$ is concatenation, and the $\max$ operation is applied on the feature dimension.
$h_{\mathcal{X}}$ is then sampled by spatial FPS \cite{fps} to guarantee equal representation to all input-segments in deeper levels of the network.
The pooling step enlarges the receptive field of the network, which is used in order to repel faraway points. The up-pooling scheme and skip-connections are similar to the ones used in PointNet++. 
An illustration of the process is depicted in figure \ref{fig:arch}.

\subsection{Soft correspondence sampling}  \label{subsec:softcor}

\CGDA presents a novel metric for ranking possible transformations and choosing the most plausible mapping between the source and target point clouds. To create the initial possible maps, we present a unique sampling method that samples efficiently the candidate points. RANSAC \cite{ransac}, which samples randomly the candidate points, requires a massive amount of possible transformations to converge to the correct solution. To overcome this, we create a smart sampling algorithm, which samples points that have a higher probability to match the target point cloud, thus inducing a procedure with much fewer required candidate transformations.

\paragraph{Soft confidence}
The soft correspondence map $\softP\in \mathbb{R}^{|\mathcal{X}|\times|\mathcal{Y}|}$ relates to the latent proximity between $x_i$ and $y_j$ and is defined by the cosine similarity between the point embeddings $\hat{h}_{x_i},\hat{h}_{y_j}$ (Eq. \ref{eq:softP}). Source points with no matched target are characterized by rows in $\softP$ with only low similarity values. This observation induces the \textit{confidence score} $\mathcal{C}$ of a point $x_i$ as the highest value in $x_i$'s row in the column of the normalized soft-alignment matrix:
\begin{equation}
\mathcal{C}_{x_i} =  \max_j \hat{\softP}_{i,j} \text{\quad where~$\hat{\softP}_{*,j}=\frac{\softP_{*,j}}{\sum\limits_{i=1}^{|\mathcal{X}|} \softP_{ij}}$}, \text{\quad}\mathcal{C}\in\mathbb{R}^{|\source|}.
\end{equation}
$\mathcal{C}_{x_i}$ is high if and only if $x_i$ is the highest correlated point to some point in $\mathcal{Y}$.
$\mathcal{C}$ is normalized to be a valid probability function $\hat{\mathcal{C}}$ which defines the sampling probability distribution $S$ over the source points that partake in the alignment. $S$ is a categorical sampling distribution over the set of source points $X = \{x_i|x_i\in\mathcal{X}\}$ with the following probability mass function:
\begin{equation}
s_{X}(x_i)=S(X=x_i)=\hat{C}_{x_i}
\end{equation}

\paragraph{Confidence guided sampling} 
\CGDnA resembles RANSAC \cite{ransac} and samples multiple possible transformations. Unlike RANSAC and its follow-up works in the registration domain, \CGDnA samples points based on the learned confidence measure of the source points in their alignment (Sec. \ref{subsec:softcor}). 
To do so, $Q$\footnote{$Q$ is a hyper-parameter and was set to $\frac{|\mathcal{X}|}{10}$ in the experiments.} points from $\source$ are sampled according to the distribution $S$, and are divided into $V=\floor{Q/r}$ experiment groups. The experiment size $r$ is constrained to $r\geq3$ as three is the minimal number of co-linear points needed to find the transformation parameters based on the SVD solution \cite{rtsolve}.

The output of the division step is $V$ experiments, where for each experiment $v$ the optimal transformation $\mathcal{T}^v$  parameters that best represent the alignment between the $r$ sampled source points $\{x_1,x_2,\dotsc,x_r\}$, and their corresponding points  $\{\pi(x_1),\pi(x_2),\dotsc,\pi(x_r)\}$ are calculated using the SVD algorithm \cite{rtsolve}. The mapping $\pi$ is set to be the maximum likelihood solution derived from $\softP$. To identify the best $\trans^v$, we use \CGDA (Sec. \ref{sec:DCGD}), and set:
\begin{equation}
    \mathcal{T}^{opt} =\argmin_{\mathcal{T}^v} CGD(\mathcal{T}^v(\source),\target).
\end{equation}
The system's novelty relies upon the power to not only suggest a transformation but also assess its correctness in a learnable manner, allowing the consensus stage to rank different suggestions.
We provide evidence for \CGDnA sampling strategy in the ablation study (Sec. \ref{Tab:ablation}), where we show a substantial performance decrease when using RANSAC.

\subsection{Self-Supervised similarity learning} \label{subsec:loss}
\CGDnA objective function is comprised of three different loss terms. The repulsion loss role is to assign each cluster of points with a different embedding space, the similarity loss encourages close embeddings for neighbor points, and the contrastive loss motivates points that appear both in the source $\source$ and target $\target$ to have the same embeddings.

\paragraph{Focal repulsion loss}
A consequential flaw of most prior works is features locality, causing spatially far co-occurring segments to have similar latent representation, yielding wrong correspondences and bad transformation solutions.
\CGDnA constrains features of spatially far points to have dissimilar features through a metric loss on the sampled points after each FPS/interpolation step (Fig. \ref{fig:arch}). Given $n_l$ pooled points, \CGDnA repulsion loss takes the form of:
\begin{equation}
    \mathcal{L}_{R_l} = \sum\limits_{i=0}^{n_l-1} \sum\limits_{j=0,j\neq i}^{n_l-1}  d(x_i,x_j)(\cos(h^l_{x_i},h^l_{x_j}))^\beta
\end{equation}
where $h^l_i$ is the embedding $h$ of point $x_i$ at layer $l$, $\cos(\cdot)$ refers to the cosine similarity, as in equation \ref{eq:cos}, and $d(x_i,x_j)$ is the Euclidean distance between the points. $\beta$ is a hyper-parameter inspired by the focal loss \cite{focal}, whose purpose is to focus on points with higher metric similarity. The function of $d(x_i,x_j)$ is to have higher weight on distant pairs.
$\mathcal{L}_{R_l}$ obliges the network to span the embeddings space and describe far points differently, even if they have locally similar geometric attributes. The unified repulsion loss is the sum of the repulsion terms, where each term is being multiplied by its index to adaptively constraint the features:
\begin{equation}
    \mathcal{L}_{R} = \sum\limits_{l=1}^{|L|}g(l) \mathcal{L}_{R_l}
\end{equation}
where $|L|$ is the number of downsampling/interpolate layers, and $g(l)$ is an increasing function putting more weight to higher layers\footnote{In practice we use $g(l)=l$}.

\paragraph{Similarity loss} For $h_{x_i}$, the final embedding, an extension of the repulsion loss is used to drive close points to have similar embeddings. The similarity loss $\mathcal{L}_{Sim}$ strives to create a smooth embedding space, which in turn increase the probability of smooth correspondence maps. Given the neighborhood $\mathcal{N}_{x_i}$ of $x_i$, the similarity loss takes the form of:
\begin{align}
    \mathcal{L}_{Sim} =& \sum\limits_{i=0}^{|\mathcal{X}|-1} \sum\limits_{j\not\in \mathcal{N}_{x_i}}  d(x_i,x_j)(\cos(h_{x_i},h_{x_j}))^\beta + \\[1pt]\nonumber  &\sum\limits_{i=0}^{|\mathcal{X}|-1} \sum\limits_{j\in \mathcal{N}_{x_i}}  \frac{1}{\max(d(x_i,x_j),\epsilon)}(1 - \cos(h_{x_i},h_{x_j}))^\beta
\end{align}
where $1 - \cos(h_{x_i},h_{x_j})$ pushes close points to have a cosine similarity close to $1$.
The RHS is divided by $d$ to emphasize the similarity between close points.

\paragraph{Contrastive Loss}
During \textbf{training}, the target shape is predefined as $\mathcal{Y} = \mathcal{T}(\mathcal{X})$ where $\mathcal{T}$ is a composition of rotation, translation and Gaussian noise. This is of course true only during training as during test different scans of the same object are used. Such target formulation defines the dense mapping during training to be  $\pi(x_i)=y_i$. Our results imply that such setting is a good proxy for the real task. \CGDnA follows DWC and uses a contrastive loss that strives for rotation and translation invariant features. The metric loss used is:
\begin{align}
    \mathcal{L}_{C} =& \sum\limits_{i=0}^{|\mathcal{X}|-1} \sum\limits_{j\not\in \mathcal{N}_{y_i}}  \cos(h_{x_i},h_{y_j}) +  \\[1pt]\nonumber &\sum\limits_{i=0}^{|\mathcal{X}|-1} \sum\limits_{j\in \mathcal{N}_{y_i}}  1 - \cos(h_{x_i},h_{y_j}).
\end{align}

\CGDnA unified loss is 
\begin{equation}\label{eq:full_loss}
    \mathcal{L} = \lambda_R\mathcal{L}_R+\lambda_{Sim}\mathcal{L}_{Sim}+\lambda_{C}\mathcal{L}_C.
\end{equation}
Ablation on the significance of the different loss functions appears in the ablation study \ref{subsec:ablation}.

\section{Experimentation}\label{sec:results}

\begin{table*}[t!]

\centering

\begin{tabular}{l@{~}c@{~}c@{~}cc@{~}c@{~}cc@{~}c@{~}cc@{~}c@{~}}
\toprule
& \multicolumn{2}{c}{Unseen point clouds}&& \multicolumn{2}{c}{Unseen categories}&& \multicolumn{2}{c}{Gaussian noise}&&
\multicolumn{2}{c}{FAUST}\\
\cmidrule(lr){2-3}
\cmidrule(lr){5-6}
\cmidrule(lr){8-9}
\cmidrule(lr){11-12}
\textbf{Model}  &\textbf{RMSE($R$)}&\textbf{RMSE($t$)}
&&\textbf{RMSE($R$)}&\textbf{RMSE($t$)}
&&\textbf{RMSE($R$)}&\textbf{RMSE($t$)}
&&\textbf{RMSE($R$)}&\textbf{RMSE($t$)}
\\
\midrule
ICP 
&24.39&0.78
&&26.35&0.85
&&27.42&0.89
&&21.92&0.69\\
 
Go-ICP 
&22.54&0.82
&&23.82&0.86
&&24.55&0.85
&&19.17&0.73\\

\noalign{\vskip 1.5mm}  
  \bottomrule 
  \noalign{\vskip 1.5mm} 
 
DCP 
&12.36&0.67
&&13.86&0.61
&&14.73&0.65
&&13.38&0.64\\

PointNetLK 
&12.12&0.55
&&14.67&0.59
&&14.98&0.61
&&11.01&0.66\\
 
IT-Net
&15.32&0.59
&&16.70&0.61
&&19.13&0.66
&&15.22&0.65\\

DWC
&19.07&0.53
&&19.26&0.61
&&20.08&0.67
&&19.08&0.41\\

PRNet
&5.93&0.32
&&6.01&0.38
&&6.93&0.41
&&6.32&0.34\\

RPM-net 
&4.93&0.1
&&5.21&0.21
&&5.09&0.25
&&11.35&0.22\\

\noalign{\vskip 1.5mm}  
  \bottomrule 
  \noalign{\vskip 1.5mm} 
  
\CGDnA (ours)
&\textbf{2.90}&\textbf{0.09}
&&\textbf{3.12}&\textbf{0.17}
&&\textbf{3.37}&\textbf{0.19}
&&\textbf{3.81}&\textbf{0.15}\\

\bottomrule
\end{tabular}
\caption{Partial-to-partial evaluations.
\CGDnA shows a considerable performance gain in all evaluation metrics on ModelNet40 (first three evaluations) and FAUST datasets, both for $R$ and $t$.}

\label{Tab:allresults}
\end{table*}
\CGDnA performance is compared to numerous learnable methods as DCP, PRNet, RPM-net, and DWC, as well as to axiomatic methods as ICP and Go-ICP. We evaluate the registration capability on multiple datasets as ModelNet40 \cite{modelnet}, Stanford Bunny \cite{stanford}, 3DMATCH \cite{3dmatch} and FAUST scans \cite{faust}. The discrepancy between the predicted $R,t$ and the ground truth transformations is measured using the root mean squared error (RMSE) metric. We measure the rotation difference using Euler angels and report the score in units of degrees. We train and evaluate \CGDnA on a single RTX8000 GPU.

\subsection{Implementation details}\label{subsec:implementation_details}
The hierarchical feature extraction network consists of two downsampling modules, with a FPS factor of 0.5, 0.25 respectively, leaving 125 points in the bottleneck of the network. The per-point output feature after the last interpolation step is concatenated with the initial $\source_{RI}$ features, and passes through 3 more $EC$ layers. The output feature dimension is 1024.
The network consists of 8.9 million learnable parameters, resulting in an architecture of size $\sim$ 15.4MB in size.  We use LeakyRelu activation function \cite{leakyrelu} with a negative slope of 0.2, and NormLayer normalization \cite{layernorm} for all the layers. 

The initial learning rate is set to $5e^{-4}$ with a multiplicative scheduler of $\gamma=0.9$ every 10 epochs. The entire training takes up to 50 epochs for the longest configuration.
For the transformation solver \cite{rtsolve}, we use the differentiable $SVD$ layer provided by PyTorch \cite{pytorch}, where we take advantage of batch-parallelization and reshape the sampled points such that each subsampled set of $k$ points acts as a new sample in a batch.
This provides two orders of magnitude speed-up compared to naively evaluating the $SVD$ per subsample, and, as we solve the least-squares problem for $k$ points, instead of $kl$ as previous methods do, our $SVD$ solver is up to 10 times faster than other methods using the $SVD$ solver.
\\ 
To create the partial view of a shape a 3D viewpoint is randomly sampled, defining a 2.5D partial shape with 60\% points of the original point cloud. Such partiality leads to pairs with $<40\%$ overlapping ratio. 
A translation $t$ sampled from $[-0.5,0.5]$ and rotation $R$ bounded to $[0\degree,60\degree]$ in each axis define the random augmentations applied separately on the source and target shapes.
We choose these augmentations to ease comparison to previous methods that limit the transformation by these parameters. However, most real-world applications encounter more extreme rotations, as in the case of multi-view registration for example.  

\subsection{ModelNet40}\label{subsec:modelnet40}

ModelNet40 \cite{modelnet} consists of 12,411 synthetic CAD models\footnote{Partial models that are globally symmetric around the axes have been filtered out. } from various categories, such as planes, furniture, cars, etc. While ModelNet40 offers the per-point normal information, \CGDnA uses only the spatial location as input information. We follow previous state-of-the-art methods and conduct 3 experiments on the ModelNet40 dataset: (1) Random train/test split - The official train-test split of ModelNet40, where 9843 samples are defined as train samples (80\% of the dataset), and the rest 2,568 samples are the test set.
The results for this test configuration are presented in table \ref{Tab:allresults}. The relative improvement compared to RPM-net, the second-best model is 2.03, where compared to DWC, the margin is higher than 16.07.  We associate the bad performance of DWC on the partial-to-partial registration problem to Chamfer's inability to "ignore" points with no matched target, contrary to \CGDA. (2) Unseen categories - In this setting we train on the first 20 categories and test on the rest. This offers an evaluation of the generalization ability of the methods to unseen topologies. The trend is similar to the previous experiment, where \CGDnA provides superior results, with sizeable margins of 2.09 $R$ RMSE compared to the second-best model (RPM-net). (3) Random noise - Noise resilience is crucial for 3D registration systems, as 3D scans from real-world sensors are known to be unstable under various factors. The noise durability test is evaluated by adding random Gaussian noise to the point clouds from the distribution $\mathcal{N}(0,0.05)$. 
Same as in the previous two test configurations, \CGDnA has the best results, where the $R$ RMSE difference is 16.71 compared to DWC and 1.72 compared to RPM-net. We ascribe this to our sampling scheme that selects an optimal \CGD solution from the experiment groups and results in outlier-free shape registration.

\subsection{FUAST scans}

The FAUST scans dataset \cite{faust} contains 300 real human scans, acquired by a real-world scanner (3dMD). The scans are partial by nature (occlusions), noisy, and contain a high number of outliers, as opposed to the ModelNet40 dataset. we do not train on FAUST scans and use the best-performing model \footnote{RNe \cite{normal1} is used as normal approximation method for RPM-net which requires normal-to-the surface.}
of each test case on ModelNet40 as the evaluated network. We do not train on real-world data as it is usually expensive to acquire and limited in size. Table \ref{Tab:allresults} provides the results on FAUST scans for the partial-to-partial task by the different models.
The results are consistent with previous tests, where \CGDnA has the best results for all the evaluated metrics. We hold \CGDnA repulsion loss (\ref{subsec:loss}) responsible for the performance boost, as human body models have many co-occurring segments (hands, palms, legs) which must have unique embeddings for a successful alignment.

\begin{figure*}[ht]
\centering

\includegraphics[width=0.67\textwidth]{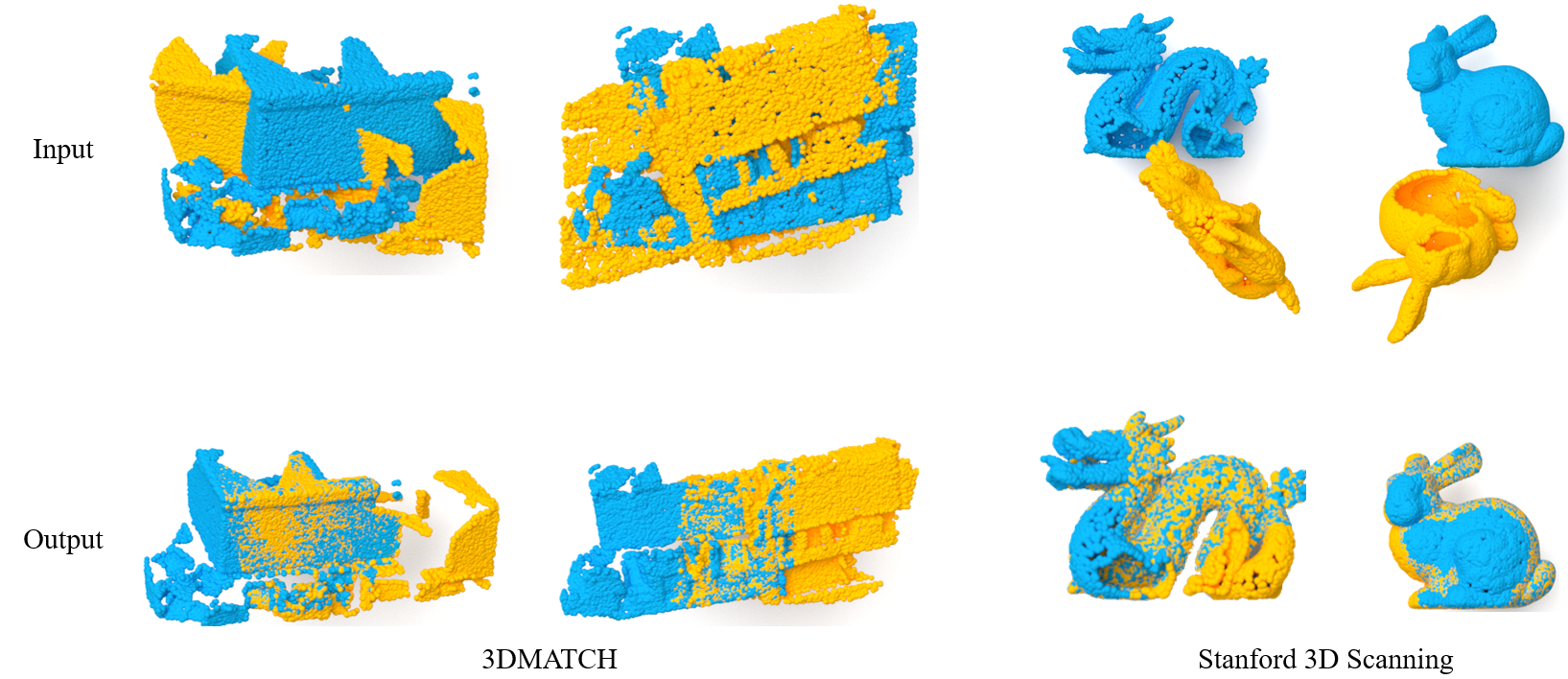}

\caption{ \textbf{Generalization capability}. Visual results of \CGDnA on 3DMATCH (left) and Stanford 3D Scanning (right). 
} \label{fig:3dmatch_bunny}
\end{figure*}

\subsection{Generalization on different modalities}\label{subsec:3dmatch}
We provide qualitative results on the Stanford 3D Scanning, and 3DMATCH datasets in figure \ref{fig:3dmatch_bunny}. While we follow monumental works as DCP, PRNet, PointNetLK, or RPM-net and focus on 3D objects, we present visual results on 2.5D indoor scans from the 3DMATCH dataset.
3DMATCH samples are noisy, and vary dramatically from ModelNet in their statistical attributes. Nevertheless, a \CGDnA trained on the \textbf{synthetic} ModelNet40 was able to extract correct registrations on this \textbf{real-world} dataset. The Stanford 3D Scanning dataset is also visually and statistically different from ModelNet40 dataset. Yet, \CGDnA offers great abilities here as well.

\subsection{Ablation Studies}\label{subsec:ablation}

\begin{table}
\centering
\caption{Ablation study on ModelNet40. The "Mode" column states the switched \textbf{off} component.
}
\label{Tab:ablation}
\begin{tabular}{l@{\hspace{0.1em}}lwc{1.2cm}p{0.1cm}wc{1.2cm}}
\toprule
&& ModelNet40&& FAUST\\
\cmidrule(lr){3-3}
\cmidrule(lr){5-5}
&\textbf{Mode}  &\textbf{RMSE($R$)}
&&
\textbf{RMSE($R$)}

\\
\midrule
(i)&\CGDA metric
&11.53&&9.52\\

(ii)&Repulsion loss
&6.18&&5.45\\
 
(iii)&Similarity loss
&5.39&&5.13\\

(iv)&Metric loss
&6.79&&6.14\\
 
(v)&Hierarchical network
&6.32&&5.88\\

&Full method
&\textbf{2.90}&&\textbf{3.81}\\
 
\bottomrule
\end{tabular}

\end{table} 
The following contributions have been evaluated in the ablation studies:
(i) \CGDA metric  - Use Chamfer distance (Eq. \ref{eq:naive_chamfer}) as the similarity metric for each experiment instead of \CGDA.
(ii) Repulsion loss - Setting $\lambda_{R}=0$ in equation \ref{eq:full_loss}, ignoring the hierarchical repulsion.
(iii) Similarity loss - Setting $\lambda_{Sim}=0$ in equation \ref{eq:full_loss}, ignoring the similarity loss.
(iv) Metric loss - Replace equation \ref{eq:full_loss} with a loss directly optimizing the transformation parameters:
\begin{equation}
    \mathcal{L} = ||R^T_{\mathcal{X}\mathcal{Y}}R^{gt}_{\mathcal{X}\mathcal{Y}} - I||^2 + ||t_{\mathcal{X}\mathcal{Y}} - t^{gt}_{\mathcal{X}\mathcal{Y}}||^2.
\end{equation}

(v) Hierarchical network - Use DGCNN \cite{dgcnn} backbone instead of PointNet++. As a result, the repulsion loss is discarded as well.

The use of \CGDA metric  provides the largest performance gain, improving the $R$ RMSE by 8.63. This affirms our hypothesis that Chamfer distance as a metric to assess partial shapes similarity is irrelevant. The gain of using the repulsion loss is also substantial, providing 3.28 improvement in $R$ RMSE. We associate it with the ability of \CGDnA to identify co-occurring segments, creating unique embeddings, and resulting in improved alignments. 

\section{Limitations}
Several design choices of \CGDnA are starting points for future work. One limitation of \CGDnA is the use of the dense alignment matrix $\softP$, as the dimensions of such map are $|\mathcal{X}|\times|\mathcal{Y}|$. Using such a soft map is sometimes unfeasible, as in the case of depth sensors for autonomous driving, acquiring $\approx 100,000$ points per scene. Accordingly, one interesting future work is to produce a hierarchical soft alignment scheme, where $\softP$ is evaluated only for sampled clusters, and propagated downwards to the dense point cloud. 
\section{Summary}\label{sec:conclusion}
In this work we examine the current state of partial-to-partial 3D rigid correspondence, particularly for large rotations, a high number of outliers, or when the input shapes have similar co-occurring segments, which are typical scenarios in real-world applications.
We present \CGDA, a new learning-based measure that fuses latent and spatial information to evaluate the similarity of point clouds and solves previous methods' inability to approximate partial point-cloud similarity.
\CGDA uses features extracted by \CGDnA, a hierarchical architecture that samples cluster centers and constrains them to have distant embeddings via the presented repulsion loss, forcing remote co-occurring segments to have unique embeddings.

\section*{Acknowledgment}
This work is partially funded by the Zimin Institute for Engineering Solutions Advancing Better Lives, the Israeli consortiums for soft robotics and autonomous driving, the Nicholas and Elizabeth Slezak Super Center for Cardiac Research and Biomedical Engineering at Tel Aviv University and TAU Science Data and AI Center.

{\small
    \bibliography{refs}

\begin{thebibliography}{42}
\providecommand{\natexlab}[1]{#1}

\bibitem[{Aoki et~al.(2019)Aoki, Goforth, Srivatsan, and Lucey}]{pointnetlk}
Aoki, Y.; Goforth, H.; Srivatsan, R.~A.; and Lucey, S. 2019.
\newblock Pointnetlk: Robust \& efficient point cloud registration using
  pointnet.
\newblock In \emph{Proceedings of the IEEE/CVF Conference on Computer Vision
  and Pattern Recognition}, 7163--7172.

\bibitem[{Ba, Kiros, and Hinton(2016)}]{layernorm}
Ba, J.~L.; Kiros, J.~R.; and Hinton, G.~E. 2016.
\newblock Layer normalization.
\newblock \emph{arXiv preprint arXiv:1607.06450}.

\bibitem[{Barrow et~al.(1977{\natexlab{a}})Barrow, Tenenbaum, Bolles, and
  Wolf}]{chamfer}
Barrow, H.~G.; Tenenbaum, J.~M.; Bolles, R.~C.; and Wolf, H.~C.
  1977{\natexlab{a}}.
\newblock Parametric correspondence and chamfer matching: Two new techniques
  for image matching.
\newblock Technical report, SRI INTERNATIONAL MENLO PARK CA ARTIFICIAL
  INTELLIGENCE CENTER.

\bibitem[{Barrow et~al.(1977{\natexlab{b}})Barrow, Tenenbaum, Bolles, and
  Wolf}]{badchamf1}
Barrow, H.~G.; Tenenbaum, J.~M.; Bolles, R.~C.; and Wolf, H.~C.
  1977{\natexlab{b}}.
\newblock Parametric correspondence and chamfer matching: Two new techniques
  for image matching.
\newblock Technical report, SRI INTERNATIONAL MENLO PARK CA ARTIFICIAL
  INTELLIGENCE CENTER.

\bibitem[{Besl and McKay(1992)}]{icp}
Besl, P.~J.; and McKay, N.~D. 1992.
\newblock Method for registration of 3-D shapes.
\newblock In \emph{Sensor fusion IV: control paradigms and data structures},
  volume 1611, 586--606. International Society for Optics and Photonics.

\bibitem[{Bogo et~al.(2014)Bogo, Romero, Loper, and Black}]{faust}
Bogo, F.; Romero, J.; Loper, M.; and Black, M.~J. 2014.
\newblock FAUST: Dataset and evaluation for 3D mesh registration.
\newblock In \emph{Proceedings of the IEEE Conference on Computer Vision and
  Pattern Recognition}, 3794--3801.

\bibitem[{Bresson et~al.(2017)Bresson, Alsayed, Yu, and
  Glaser}]{autonomous_registration}
Bresson, G.; Alsayed, Z.; Yu, L.; and Glaser, S. 2017.
\newblock Simultaneous localization and mapping: A survey of current trends in
  autonomous driving.
\newblock \emph{IEEE Transactions on Intelligent Vehicles}, 2(3): 194--220.

\bibitem[{Chen and Medioni(1992)}]{chen1992object}
Chen, Y.; and Medioni, G. 1992.
\newblock Object modelling by registration of multiple range images.
\newblock \emph{Image and vision computing}, 10(3): 145--155.

\bibitem[{Choy, Dong, and Koltun(2020)}]{deepglobal}
Choy, C.; Dong, W.; and Koltun, V. 2020.
\newblock Deep global registration.
\newblock In \emph{Proceedings of the IEEE/CVF conference on computer vision
  and pattern recognition}, 2514--2523.

\bibitem[{Dang, Wang, and Salzmann(2020)}]{3d3d}
Dang, Z.; Wang, F.; and Salzmann, M. 2020.
\newblock Learning 3D-3D Correspondences for One-shot Partial-to-partial
  Registration.
\newblock \emph{CoRR}, abs/2006.04523.

\bibitem[{Durrant-Whyte and Bailey(2006)}]{robotics_registration}
Durrant-Whyte, H.; and Bailey, T. 2006.
\newblock Simultaneous localization and mapping: part I.
\newblock \emph{IEEE robotics \& automation magazine}, 13(2): 99--110.

\bibitem[{Eldar et~al.(1997)Eldar, Lindenbaum, Porat, and Zeevi}]{fps}
Eldar, Y.; Lindenbaum, M.; Porat, M.; and Zeevi, Y. 1997.
\newblock The farthest point strategy for progressive image sampling.
\newblock \emph{IEEE transactions on image processing : a publication of the
  IEEE Signal Processing Society}, 6 9: 1305--15.

\bibitem[{Fischler and Bolles(1981)}]{ransac}
Fischler, M.~A.; and Bolles, R.~C. 1981.
\newblock Random Sample Consensus: A Paradigm for Model Fitting with
  Applications to Image Analysis and Automated Cartography.
\newblock \emph{Commun. ACM}, 24(6): 381–395.

\bibitem[{Fitzgibbon(2003)}]{icpv1}
Fitzgibbon, A.~W. 2003.
\newblock Robust registration of 2D and 3D point sets.
\newblock \emph{Image and vision computing}, 21(13-14): 1145--1153.

\bibitem[{Ginzburg and Raviv(2020)}]{cyclic}
Ginzburg, D.; and Raviv, D. 2020.
\newblock Cyclic functional mapping: Self-supervised correspondence between
  non-isometric deformable shapes.
\newblock In \emph{European Conference on Computer Vision}, 36--52. Springer.

\bibitem[{Ginzburg and Raviv(2021{\natexlab{a}})}]{dwc}
Ginzburg, D.; and Raviv, D. 2021{\natexlab{a}}.
\newblock Deep Weighted Consensus: Dense correspondence confidence maps for 3D
  shape registration.
\newblock arXiv:2105.02714.

\bibitem[{Ginzburg and Raviv(2021{\natexlab{b}})}]{DG2N}
Ginzburg, D.; and Raviv, D. 2021{\natexlab{b}}.
\newblock Dual Geometric Graph Network (DG2N) Iterative Network for Deformable
  Shape Alignment.
\newblock \emph{2021 International Conference on 3D Vision (3DV)}.

\bibitem[{Hajnal and Hill(2001)}]{medical_registration}
Hajnal, J.~V.; and Hill, D.~L. 2001.
\newblock \emph{Medical image registration}.
\newblock CRC press.

\bibitem[{Huang et~al.(2021)Huang, Gojcic, Usvyatsov, and
  Andreas~Wieser}]{predator}
Huang, S.; Gojcic, Z.; Usvyatsov, M.; and Andreas~Wieser, K.~S. 2021.
\newblock PREDATOR: Registration of 3D Point Clouds with Low Overlap.
\newblock In \emph{IEEE Conference on Computer Vision and Pattern Recognition,
  CVPR}.

\bibitem[{Kabsch(1976)}]{rtsolve}
Kabsch, W. 1976.
\newblock A solution for the best rotation to relate two sets of vectors.
\newblock \emph{Acta Crystallographica Section A: Crystal Physics, Diffraction,
  Theoretical and General Crystallography}, 32(5): 922--923.

\bibitem[{Kim et~al.(2012)Kim, Li, Mitra, DiVerdi, and Funkhouser}]{icpl3}
Kim, V.~G.; Li, W.; Mitra, N.~J.; DiVerdi, S.; and Funkhouser, T. 2012.
\newblock Exploring collections of 3d models using fuzzy correspondences.
\newblock \emph{ACM Transactions on Graphics (TOG)}, 31(4): 1--11.

\bibitem[{Lang et~al.(2021)Lang, Ginzburg, Avidan, and Raviv}]{dpc}
Lang, I.; Ginzburg, D.; Avidan, S.; and Raviv, D. 2021.
\newblock DPC: Unsupervised Deep Point Correspondence via Cross and Self
  Construction.
\newblock In \emph{2021 International Conference on 3D Vision (3DV)},
  1442--1451.

\bibitem[{Li et~al.(2010)Li, Schnabel, Klein, Cheng, Dang, and Jin}]{normal1}
Li, B.; Schnabel, R.; Klein, R.; Cheng, Z.; Dang, G.; and Jin, S. 2010.
\newblock Robust normal estimation for point clouds with sharp features.
\newblock \emph{Computers \& Graphics}, 34(2): 94--106.

\bibitem[{Li et~al.(2018)Li, Wang, Zhu, and Huang}]{gnns1}
Li, R.; Wang, S.; Zhu, F.; and Huang, J. 2018.
\newblock Adaptive graph convolutional neural networks.
\newblock In \emph{Proceedings of the AAAI Conference on Artificial
  Intelligence}, volume~32.

\bibitem[{Lin et~al.(2017)Lin, Goyal, Girshick, He, and Doll{\'{a}}r}]{focal}
Lin, T.; Goyal, P.; Girshick, R.~B.; He, K.; and Doll{\'{a}}r, P. 2017.
\newblock Focal Loss for Dense Object Detection.
\newblock \emph{CoRR}, abs/1708.02002.

\bibitem[{Paszke et~al.(2019)Paszke, Gross, Massa, Lerer, Bradbury, Chanan,
  Killeen, Lin, Gimelshein, Antiga, Desmaison, Kopf, Yang, DeVito, Raison,
  Tejani, Chilamkurthy, Steiner, Fang, Bai, and Chintala}]{pytorch}
Paszke, A.; Gross, S.; Massa, F.; Lerer, A.; Bradbury, J.; Chanan, G.; Killeen,
  T.; Lin, Z.; Gimelshein, N.; Antiga, L.; Desmaison, A.; Kopf, A.; Yang, E.;
  DeVito, Z.; Raison, M.; Tejani, A.; Chilamkurthy, S.; Steiner, B.; Fang, L.;
  Bai, J.; and Chintala, S. 2019.
\newblock PyTorch: An Imperative Style, High-Performance Deep Learning Library.
\newblock In Wallach, H.; Larochelle, H.; Beygelzimer, A.; d\textquotesingle
  Alch\'{e}-Buc, F.; Fox, E.; and Garnett, R., eds., \emph{Advances in Neural
  Information Processing Systems 32}, 8024--8035. Curran Associates, Inc.

\bibitem[{Pazi, Ginzburg, and Raviv(2020)}]{unsupmultispectral}
Pazi, I.; Ginzburg, D.; and Raviv, D. 2020.
\newblock Unsupervised Scale-Invariant Multispectral Shape Matching.
\newblock \emph{arXiv preprint arXiv:2012.10685}.

\bibitem[{Qi et~al.(2017)Qi, Yi, Su, and Guibas}]{pointnetpp}
Qi, C.~R.; Yi, L.; Su, H.; and Guibas, L.~J. 2017.
\newblock Pointnet++: Deep hierarchical feature learning on point sets in a
  metric space.
\newblock In \emph{Advances in neural information processing systems},
  5099--5108.

\bibitem[{Sharp, Lee, and Wehe(2002)}]{sharp2002icp}
Sharp, G.~C.; Lee, S.~W.; and Wehe, D.~K. 2002.
\newblock ICP registration using invariant features.
\newblock \emph{IEEE Transactions on Pattern Analysis and Machine
  Intelligence}, 24(1): 90--102.

\bibitem[{Turk and Levoy(1994)}]{stanford}
Turk, G.; and Levoy, M. 1994.
\newblock Zippered Polygon Meshes from Range Images.
\newblock In \emph{Proceedings of the 21st Annual Conference on Computer
  Graphics and Interactive Techniques}, SIGGRAPH '94, 311–318. New York, NY,
  USA: Association for Computing Machinery.
\newblock ISBN 0897916670.

\bibitem[{Wang and Solomon(2019{\natexlab{a}})}]{dcp}
Wang, Y.; and Solomon, J.~M. 2019{\natexlab{a}}.
\newblock Deep closest point: Learning representations for point cloud
  registration.
\newblock In \emph{Proceedings of the IEEE/CVF International Conference on
  Computer Vision}, 3523--3532.

\bibitem[{Wang and Solomon(2019{\natexlab{b}})}]{prnet}
Wang, Y.; and Solomon, J.~M. 2019{\natexlab{b}}.
\newblock Prnet: Self-supervised learning for partial-to-partial registration.
\newblock \emph{arXiv preprint arXiv:1910.12240}.

\bibitem[{Wang et~al.(2019)Wang, Sun, Liu, Sarma, Bronstein, and
  Solomon}]{dgcnn}
Wang, Y.; Sun, Y.; Liu, Z.; Sarma, S.~E.; Bronstein, M.~M.; and Solomon, J.~M.
  2019.
\newblock Dynamic graph cnn for learning on point clouds.
\newblock \emph{Acm Transactions On Graphics (tog)}, 38(5): 1--12.

\bibitem[{Wu et~al.(2015)Wu, Song, Khosla, Yu, Zhang, Tang, and
  Xiao}]{modelnet}
Wu, Z.; Song, S.; Khosla, A.; Yu, F.; Zhang, L.; Tang, X.; and Xiao, J. 2015.
\newblock 3d shapenets: A deep representation for volumetric shapes.
\newblock In \emph{Proceedings of the IEEE conference on computer vision and
  pattern recognition}, 1912--1920.

\bibitem[{Xu et~al.(2015)Xu, Wang, Chen, and Li}]{leakyrelu}
Xu, B.; Wang, N.; Chen, T.; and Li, M. 2015.
\newblock Empirical evaluation of rectified activations in convolutional
  network.
\newblock \emph{arXiv preprint arXiv:1505.00853}.

\bibitem[{Yan et~al.(2021)Yan, Yi, Hu, Mitra, Cohen-Or, and Huang}]{dec2}
Yan, Z.; Yi, Z.; Hu, R.; Mitra, N.; Cohen-Or, D.; and Huang, H. 2021.
\newblock Consistent Two-Flow Network for Tele-Registration of Point Clouds.
\newblock \emph{IEEE Transactions on Visualization and Computer Graphics},
  1--1.

\bibitem[{Yang et~al.(2015)Yang, Li, Campbell, and Jia}]{icpl1}
Yang, J.; Li, H.; Campbell, D.; and Jia, Y. 2015.
\newblock Go-ICP: A globally optimal solution to 3D ICP point-set registration.
\newblock \emph{IEEE transactions on pattern analysis and machine
  intelligence}, 38(11): 2241--2254.

\bibitem[{Yang, Yan, and Huang(2019)}]{dec3}
Yang, Z.; Yan, S.; and Huang, Q. 2019.
\newblock Extreme Relative Pose Network under Hybrid Representations.
\newblock \emph{CoRR}, abs/1912.11695.

\bibitem[{Yew and Lee(2020)}]{robustpointcloud}
Yew, Z.~J.; and Lee, G.~H. 2020.
\newblock Rpm-net: Robust point matching using learned features.
\newblock In \emph{Proceedings of the IEEE/CVF conference on computer vision
  and pattern recognition}, 11824--11833.

\bibitem[{Yuan et~al.(2018)Yuan, Held, Mertz, and Hebert}]{itnet}
Yuan, W.; Held, D.; Mertz, C.; and Hebert, M. 2018.
\newblock itnet.
\newblock \emph{CoRR}, abs/1811.11209.

\bibitem[{Zeng et~al.(2017)Zeng, Song, Nie{\ss}ner, Fisher, Xiao, and
  Funkhouser}]{3dmatch}
Zeng, A.; Song, S.; Nie{\ss}ner, M.; Fisher, M.; Xiao, J.; and Funkhouser, T.
  2017.
\newblock 3DMatch: Learning Local Geometric Descriptors from RGB-D
  Reconstructions.
\newblock In \emph{CVPR}.

\bibitem[{{Zinsser}, {Schmidt}, and {Niemann}(2003)}]{icpl4}
{Zinsser}, T.; {Schmidt}, J.; and {Niemann}, H. 2003.
\newblock A refined ICP algorithm for robust 3-D correspondence estimation.
\newblock In \emph{Proceedings 2003 International Conference on Image
  Processing (Cat. No.03CH37429)}, volume~2, II--695.

\end{thebibliography}
}
\newpage

\end{document}